# ExpressMM: Expressive Mobile Manipulation Behaviors in Human-Robot Interactions

Souren Pashangpour*, Haitong Wang*, *IEEE Student Member,* Matthew Lisondra*, *IEEE Student Member*, and Goldie Nejat, *IEEE Member*

*Abstract*— **Mobile manipulators are increasingly deployed in human-centered environments to perform tasks. While completing such tasks, they should also be able to communicate their intent to the people around them using expressive robot behaviors. Prior work on expressive robot behaviors has used preprogrammed or learning from demonstration- based expressive motions, and/or large language model-generated high-level interactions. The majority of these existing approaches have not considered human-robot interactions (HRI) where users may interrupt, modify, or redirect a robot's actions during task execution. In this paper, we develop the novel ExpressMM framework that integrates a high-level language-guided planner based on a vision-language model for perception and conversational reasoning, with a low-level vision-language-action policy to generate expressive robot behaviors during collaborative HRI tasks. Furthermore, ExpressMM supports interruptible interactions to accommodate updated or redirecting instructions by users. We demonstrate ExpressMM on a mobile manipulator assisting a human in a collaborative assembly scenario and conduct audience-based evaluation of live HRI demonstrations. Questionnaire results show that the ExpressMM-enabled expressive behaviors helped observers clearly interpret the robot's actions and intentions while supporting socially appropriate and understandable interactions. Participants also reported that the robot was useful for collaborative tasks and behaved in a predictable and safe manner during the demonstrations, fostering positive perceptions of the robot's usefulness, safety, and predictability during the collaborative tasks.**

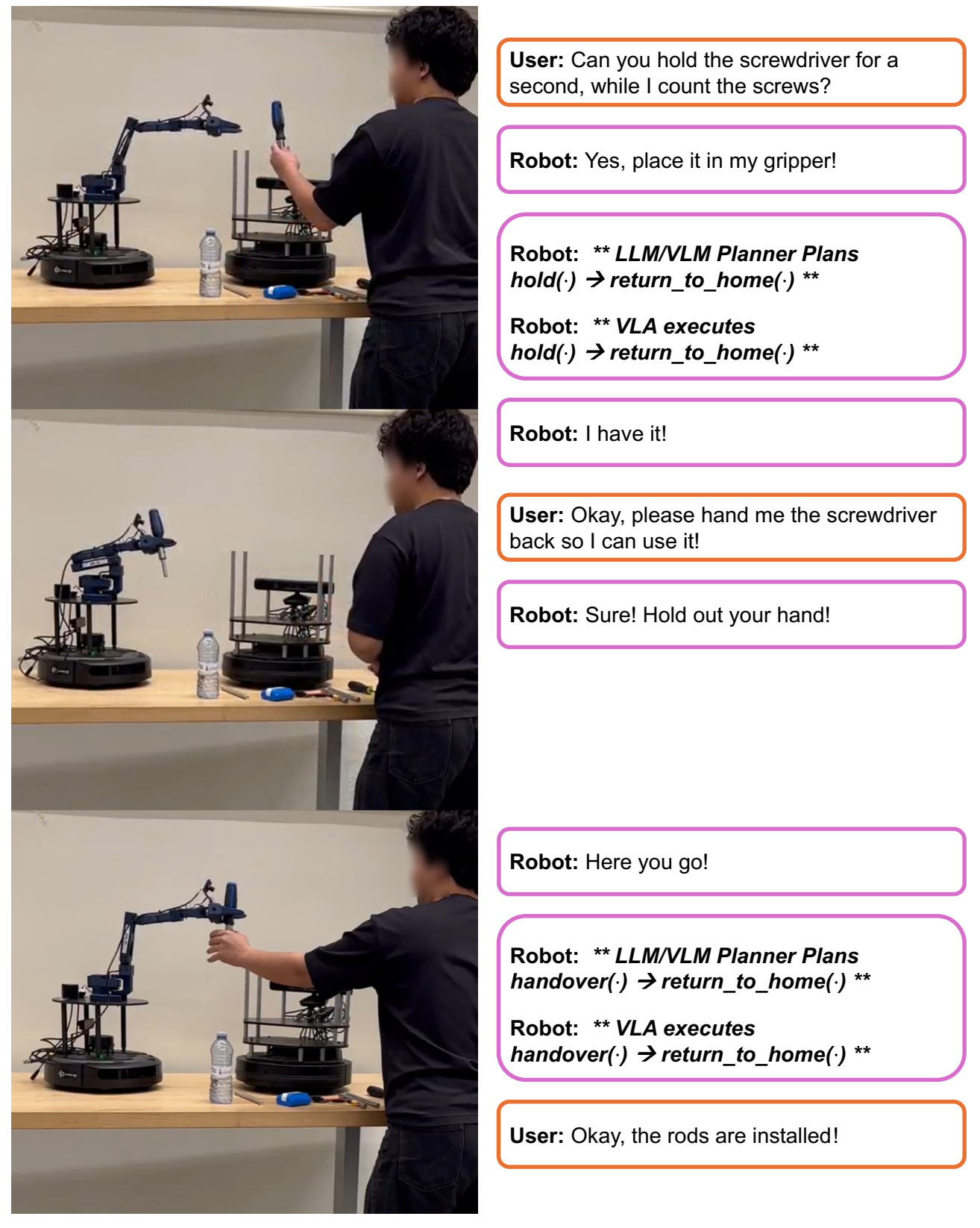

**Fig. 1.** Example HRI scenario during a collaborative assembly task using the proposed ExpressMM framework. The robot holds and returns a screwdriver through language-guided planning and VLA execution.

## I. INTRODUCTION

Mobile manipulator service robots are increasingly being incorporated in human-centered environments such as homes [1], [2] and hospitals [3], [4]. In these settings, robots work with or in close proximity to humans, performing such tasks as search [5], fetch [6], carrying [7] and handover [8] in shared spaces. Their deployment success not only depends on task completion but also on how their behaviors are perceived by human collaborators. Such mobile manipulators may not have an anthropomorphic or human-like embodiment, and need to use their movements to engage in social human-robot interactions (HRI). In particular, the robots need to be able to convey their intentions and states to humans [9].

An open challenge for mobile manipulation robots engaged in HRI is how a robot can display its intent directly to a user while still satisfying functional and safety constraints. For example, during a handover task, a single-arm mobile manipulator may need to convey "*I'm ready*," "*please take it*," or "*I'm waiting for you*," using both arm gestures and base positioning and orientation. Furthermore, while navigating human-centered environments, the robot may need to indicate "*I will pass behind you*," "*I'm yielding*," or "*excuse me*," rather than replanning its navigation path around people.

In the aforementioned scenarios, expressive motion can be utilized to help with the interpretation of robot intent and actions. Such expressive robot motion has been designed using different approaches that focus on incorporating

This work was supported by the Natural Sciences and Engineering Research Council of Canada (NSERC), the Canada Research Chair program, the Canadian Institute for Applied Research (CIFAR), and a NSERC CREATE ADVENTOR Fellowship. *Corresponding author: Souren Pashangpour.*

* Indicates that the authors contributed equally to the paper. The authors are with the Autonomous Systems and Biomechatronics Laboratory (ASBLab), Department of Mechanical and Engineering, University of Toronto, Toronto, ON M5S 3G8, Canada (e-mail: souren.pashangpour@mail.utoronto.ca; haitong.wang@mail.utoronto.ca; matthew.lisondra@mail.utoronto.ca; goldie.nejat@utoronto.ca).

communicative cues directly into robot motion generation and control. Namely, movement-centric robot frameworks have jointly incorporated expressive and functional objectives in motion design to allow both robot goal achievement and the communication of a robot's intentions, attention, and interaction stance (e.g., signaling readiness, hesitation, or invitation to interact) [10], [11], [12]. This has been primarily achieved through the manual development of motion templates. Data-driven approaches learn expressive robot motions or motion style representations from learning from demonstration (LfD) by people or other robots performing the tasks while conveying interaction cues [13], [14], [15]. This allows for the capture of task execution variations (e.g., cautious, confident, or hesitant motion trajectories). However, data-dirven methods require potentially large sets of demonstrations. Recently, foundation-models such as large language models (LLMs) or vision language models (VLMs) have been used to translate language instructions and social context into expressive robot behaviors by generating high-level behavioral plans or interaction primitives that guide robot motion generation [16], [17], [18]. These LLM-based approaches mainly focus on high-level symbolic planning or language reasoning.

To the authors' knowledge, the aforementioned methods have typically assume a fixed task instruction and are not able to support interruptible user requests during task execution. However, in collaborative mobile manipulation tasks, users may issue corrective or redirecting commands, while a robot is in the process of completing a task [19], [20], [21], [22]. Supporting such interaction requires maintaining conversational context and updating the robot's action sequence while ensuring that the resulting behaviors remain physically executable.

In this paper, we present the development of a multi-foundation model framework, ExpressMM, for expressive intent communication by mobile manipulators engaged in HRI. ExpressMM integrates a combination of a high-level language-guided planner based on a VLM, and a low-level vision-language-action (VLA) policy to enable the generation for expressive robot behaviors during collaborative tasks, Fig. 1. The novelty of ExpressMM is in its incorporation of language- and perception-driven expressive interaction behaviors that are able to handle conversational interruption updates from users. ExpressMM has been incorporated into a single-arm mobile manipulator and evaluated in an audience-based live demonstration of a collaborative human-robot task.

## II. RELATED WORKS

Prior work on expressive robot behaviors for non-anthropomorphic robots engaged in HRI can be categorized into: 1) expressive robot motion methods [10], [11], [12], [13], [14], [15], [16], [17], [18], and 2) interruptible requests for robots [19], [20], [21], [22].

### A. *Expressive Robot Motion Methods*

Expressive robot motion methods for HRI can be categorized into: 1) movement-centric [10], [11], [12], 2) data-driven [13], [14], [15], and 3) foundation-model-based approaches [16], [17], [18].

Movement-centric methods pre-program communicative intent into a robot's motion through spatial and temporal design choices, such as approach direction, pauses, speed, posture [10], [11], and the robot's form factors, such as degree-of-freedom (DOF) configurations [12]. In [10], a lamp-like non-anthropomorphic robot was designed to explore the interplay between functional and expressive movement objectives. A user study comparing expression-driven and function-driven movements showed that expression-driven movements improved user engagement and perceived robot qualities, especially in social-oriented tasks. In [11], an artificial emotional intelligence system was developed to manually encode emotion into preprogrammed poses, gestures, and motion parameters for the Shimi interactive robot. In a user study, participants viewed Shimi's emotional expressions as either static postures or dynamic gestures, presented on a screen or in-person. The results showed that dynamic gestures conveyed emotions more accurately than static postures. In [12], expressive movements were developed by first exploring gestures and different DOF configurations through 3D animation studies and then tested on a physical robot skeleton. Lastly, video recordings of the robot performing different motions were shown. The results showed that compared with a screen-only setup, the added robot motions improved behavior understanding, increased confidence in interpretation, and increased the perceived involvement of a remote user.

Data-driven methods use LfD [13], [14], [15] to learn expressive robot motions. Namely, these methods learn the mapping from demonstrated motion sequences or labeled preferences to affective or social movement styles. For example, in [13], a variational autoencoder was trained on hand-crafted emotive movement examples collected on the Blossom interactive robot in order to modify existing motions into different emotional styles. In an online survey, participants watched videos of original and modified Blossom motions. The results showed that the modified motions remained as legible as the original motions for all target emotions. In [14], two style cost functions were learned for a robot manipulator to generate expressive motions. The functions consisted of: 1) a linear combination of hand-designed features, and 2) a neural network on raw trajectories. They were learned from human preference labels on candidate motions generated by trajectory optimization. In a user study, participants rated carry, place, and handover motions generated for the robot manipulator in happy, sad, and hesitant styles. Results showed both learned cost functions produced robot motions that were rated more expressive than a baseline method without learned style cost functions. In [15], a style discriminator was interactively trained to map robot trajectories to a continuous valence-arousal-dominance (VAD) emotion space by learning from human VAD labels. This learned mapping was then converted into a style cost, so a trajectory optimizer could generate task motions that both completed the task and matched a target emotion. Experiments with a simulated vacuum robot and the Cassie biped robot showed that learning a single style discriminator over the continuous VAD space was more efficient than emotion-specific cost functions. In a user study, users labeled robot trajectories with VAD scores and during evaluation the generated motions were recognized higher than a random-guess across target emotions.

More recently, foundation-model approaches have used LLMs or VLMs to map natural-language interaction instructions and social context for expressive robot control into robot high-level action sequences [16], [17], [18]. For example, in [16], the Generative Expressive Motion (GenEM) architecture was developed to use few-shot chain-of-thought prompting with GPT-4 [23] to translate natural-language instructions of a desired expressive behavior into robot behaviors by generating a human-like response description, and robot behaviors. In online user studies using videos of a mobile robot performing expressive HRI behaviors such as nodding and approaching, the generated behaviors were rated by participants as more competent and easier to understand, and were comparable to behaviors designed by a professional animator. In [17], the Language-Informed Latent Actions with Corrections (LILAC) framework was developed in order for a user to guide a robot arm using language during manipulation tasks. The framework used a frozen Distil-RoBERTa [24] language encoder and a GPT-3 [25] gating module to combine the robot state, the initial task instruction, the user's input and corrections to update the robot's motion online. In a user study with a Franka Emika Panda manipulator, LILAC achieved the highest success rate across all tasks, and was preferred over baseline learning methods. In [18], an LLM-based framework was developed to map target robot states to expressive quadruped robot motions. It used GPT-4 [23] to select a quadruped motion from a fixed set of predefined motions (body orientation and velocity). In two online studies, half of the participants proposed robot states and selected motion parameters, and another half watched videos of the quadruped expressing states under LLM-generated, human-selected, and random-motion conditions to classify the robot's state. The LLM-generated motions outperformed the random baseline and were comparable to human-selected motions.

## B. *Interruptible Requests for Robots*

Mobile manipulation in human-centered environments requires robot behaviors to remain interruptible during execution, since users frequently provide mid-task corrections [20] and the changing environments may invalidate previously feasible plans [19]. In general, interruption-aware systems monitor the robot's ongoing task, user input, and environment state during execution, and when a new command or change is detected, they stop or modify the affected action and update the task or motion plan so that the robot can continue safely [19], [20], [21], [22]. For example, in [19], a reactive task and motion planning framework was developed to update the portion of an object-transfer plan affected by a human intervention, and reconfigure a behavior tree online so task execution could continue without stopping and replanning from scratch. In object transfer experiments with a UR5 robot, the method achieved lower replanning and task completion time compared to A* [26] and Dijkstra [27] baselines. In [20], an interruptible-autonomy framework was presented for a mobile service robot. The framework detected an interruption when a person stopped the robot and started speech interactions. It then parsed the spoken command into actions such as status inquiry, task cancellation, or scheduling a new task, and then rescheduled the robot's task list. In demonstrations with a CoBot mobile service robot, the system allowed users to query the robot's current task, cancel an active task, insert a new task for immediate execution, or schedule a new task after the current one was completed.

In [21], a correction-handling architecture was developed to allow uasers to interrupt and revise spoken robot instructions during navigation, retrieval, and meal-assembly tasks. The architecture detected corrective utterances, replaced the original goal or action, and generated undo steps when needed before continuing with the updated task. In an online user study, participants were shown scenes with blocks, tools, or map locations, together with an initial instruction and a correction. They were asked to select the object or destination intended by the corrected instruction; the framework matched human interpretations in the majority of cases. In [22], the Hi Robot system was developed to enable interruptible open-ended robot manipulation, where the robot interpreted complex task instructions, user feedback, and mid-task corrections during execution. The system consisted of a high-level VLM and a low-level VLA where the VLM was rerun immediately when the user provided a new spoken or text correction. In experiments, the system handled mid-task corrections and additional requests during execution, and outperformed GPT-4o high-level and flat VLA baselines in instruction accuracy and task progress.

## C. *Summary of Limitations*

Movement-centric approaches are often embodiment-specific and difficult to generalize to new mobile manipulation tasks [10], [11], [12]. Data-driven methods depend on curated training data and robot-specific motion parameterizations [13], [14], [15]. Foundation-model-based approaches often rely on predefined low-level action primitives or motion parameters [16], [17], [18]. As a result, extending them to new skills, tasks, or environments often requires additional manual design of new action primitives. Interruption-aware execution methods have not included expressive motion generation [19], [20], [21], [22]. They update a robot's task after interruption and they do not consider how a robot should communicate intent during the interaction.

To address the above limitations, we have developed ExpressMM, an expressive, interruptible mobile manipulation framework that utilizes the generalization capbilities of a VLM and VLA to generate online expressive robot motions, without relying on fixed motion primitives or curated demonstration datasets. ExpressMM integrates conversational interruption handling within the planning architecture, enabling the robot to update expressive interaction actions in response to user corrections during task execution while maintaining coordinated whole-body motion generation.

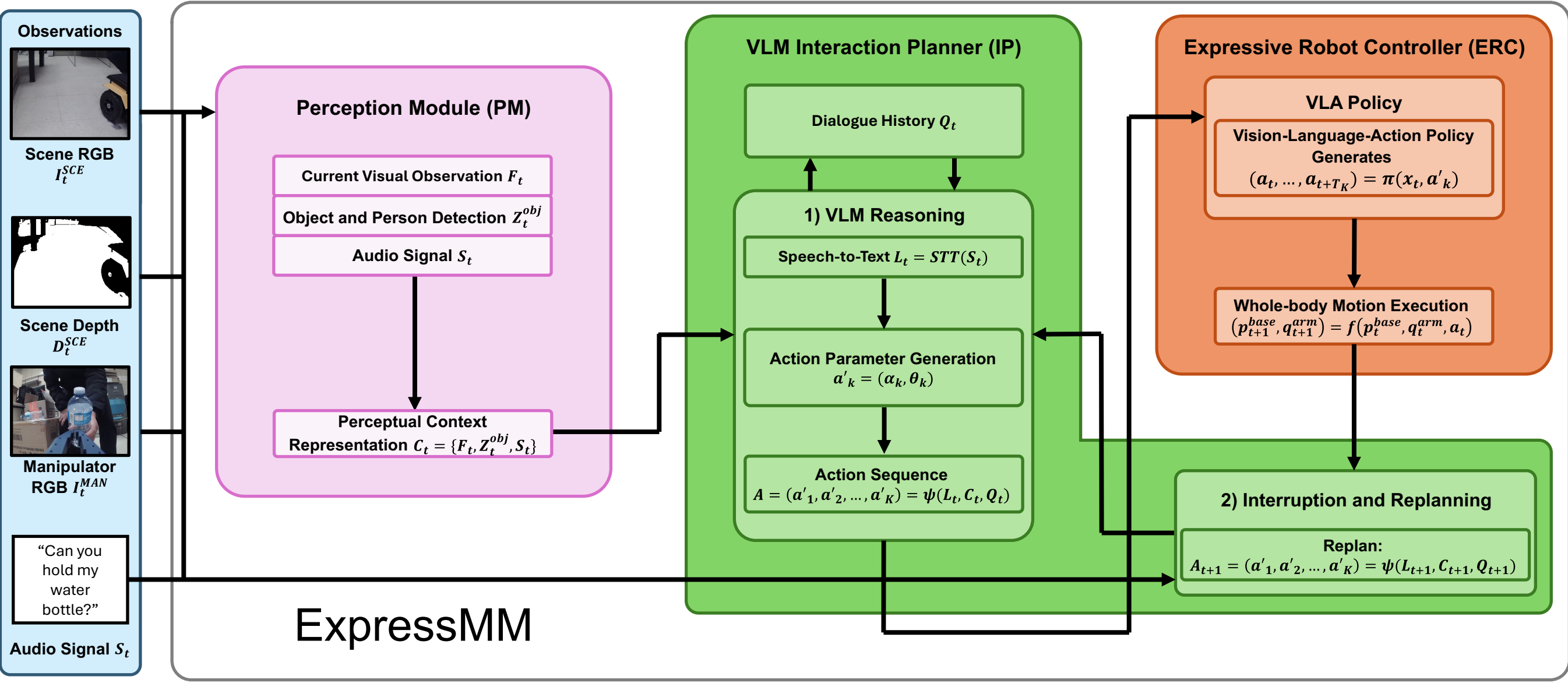

**Fig. 2.** The proposed expressive mobile manipulation architecture consisting of: 1) a Perception Module (PM), 2) a VLM Interaction Planner (IP), and 3) an Expressive Robot Controller (ERC).

## III. EXPRESSIVE MOBILE MANIPULATION ARCHITECTURE

The proposed architecture for the ExpressMM consists of three main modules, Fig. 2: 1) Perception Module (PM), 2) VLM Interaction Planner (IP), and 3) Expressive Robot Controller (ERC).

Multimodal sensory observations are provided to the PM to estimate the current scene state and infer interaction cues from the user. The PM uses both visual and audio observations to generate a structured scene representation containing detected objects, human pose information, and interpreted user interaction signals. This scene representation, together with the task instruction obtained from the user's speech input, is provided to the interaction planner. The VLM IP generates a behavior plan represented as a sequence of expressive robot action primitives, where each primitive specifies a task-relevant interaction behavior (e.g., approach, present, wait, retreat) together with associated expressive motion parameters. The generated action sequence is then provided to the ERC, which executes each action using a VLA policy to generate coordinated robot motions.

### *A. Perception Module (PM)*

The Perception Module uses multimodal sensory observations from the robot and provides perceptual inputs required for interaction reasoning and action execution. At timestep $t$, the robot receives multimodal observations $O_t = (I_t^{SCE}, D_t^{SCE}, I_t^{MAN}, S_t)$ where $I_t^{SCE}$ denotes the visual observation obtained from an RGB image of the surrounding environment; this image contains visual information about nearby humans and objects and is used during navigation and interaction. $D_t^{SCE}$ denotes the corresponding depth observation which provides the geometric information about the scene. $I_t^{MAN}$ denotes an RGB image of the robot manipulation workspace. $S_t$ is the user audio signal captured by the robot microphone.

Using the aforementioned observations, a perceptual context representation $C_t = \{F_t, Z_t^{obj}, S_t\}$ is obtained, where $F_t$ denotes the current visual observation used as visual context for the vision-language planner and $Z_t^{obj}$ represents the set of detected task-relevant objects/persons in the scene.

Object and person instances $Z_t^{obj}$ are detected by applying YOLOv8-nano [28] to the RGB observation $I_t^{SCE}$, producing bounding boxes $B = \{b_1, b_2, \dots, b_m\}$ where $b_i$ denotes the bounding box corresponding to detected instance $i$. Depth observations are used to estimate the 3D centroid of each detected object to determine the spatial location of objects relative to the robot.

The resulting perceptual context $C_t$ is provided to the VLM IP in order to combine the user's spoken instructions with the visual observations of the scene to determine the appropriate expressive robot actions.

### *B. VLM Interaction Planner (IP)*

The VLM Interaction Planner determines a sequence of expressive robot actions that both accomplish the requested task and communicate the robot's intended interaction behavior to the user.

*1) VLM Reasoning:* The planner receives as input $(L_t, C_t, Q_t)$, where $L_t$ denotes the current natural-language instruction obtained from user speech; the instruction $L_t$ is obtained by converting the audio signal $M_t$ into text using a speech recognition submodule $L_t = STT(S_t)$ where $STT(\cdot)$ denotes the speech-to-text function. $Q_t$ denotes the current full session dialogue history (i.e. since initialization) between the user and the robot.

The planner generates expressive actions using VLM GPT-5.2 [29] which maps the current interaction context to a sequence of expressive robot actions:

$$A = (a'_1, a'_2, \dots, a'_K) = \psi(L_t, C_t, Q_t), \quad (1)$$

where each action, $a'_k = (\alpha_k, \theta_k)$ corresponds to an interaction behavior parameterized by motion intent. $\alpha_k$ denotes the action type (e.g. approach, present, wait, retreat, follow, handover); $\theta_k$ denotes the expressive motion parameters associated with the action, such as gesture style, motion intensity, or robot orientation relative to the user; and $\psi(\cdot)$ denotes the VLM GPT-5.2 [29] interaction planner. The planner integrates language reasoning and visual scene interpretation from the VLM GPT-5.2 [29] to jointly interpret the spoken user instruction, the perceived scene context, and the dialogue history when selecting expressive robot actions.

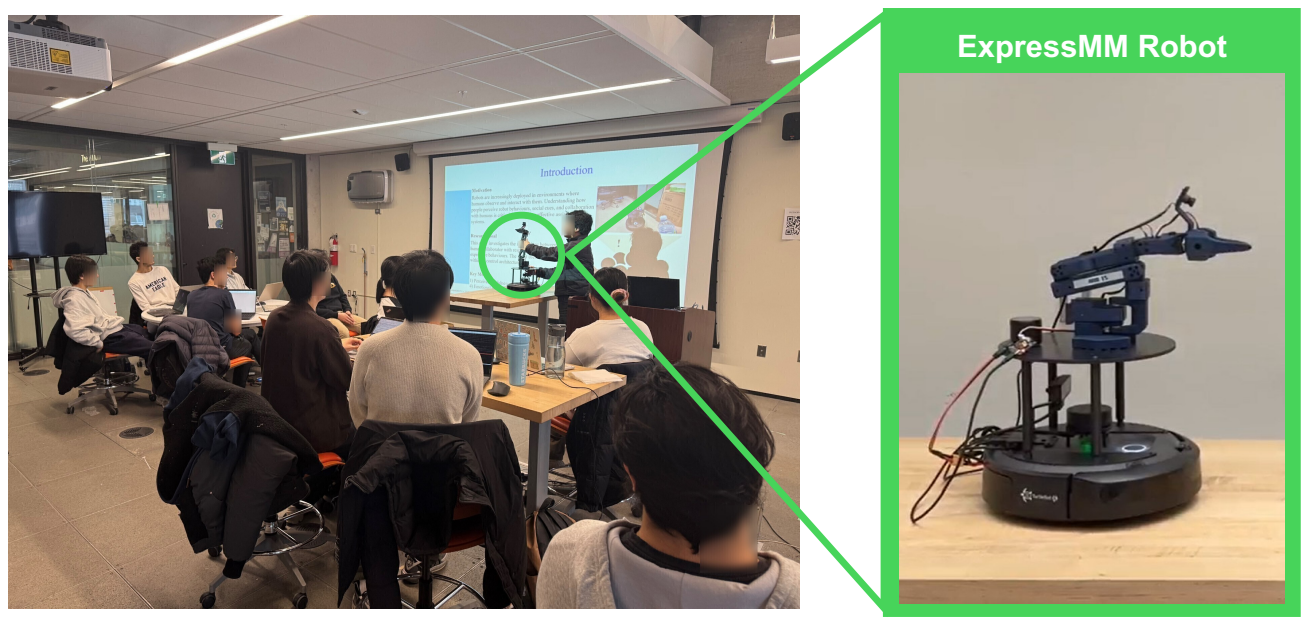

**Fig. 3.** Setup of the Turtlebot-4 with SO-101 arm mobile manipulator robot interacting with a researcher in front of a classroom audience.

The resulting expressive action sequence $A$ defines the expressive interaction plan and is provided to the ERC for execution.

*2) Interruption and Replanning:* If a new instruction $L_{t+1}$ is detected or the perceptual context updates $C_{t+1}$ due to new observations, the interaction state becomes $(L_{t+1}, C_{t+1}, Q_{t+1})$. The planner then recomputes the expressive interaction plan using Eq. 1:

$$A_{t+1} = (a'_1, a'_2, \dots, a'_K) = \psi(L_{t+1}, C_{t+1}, Q_{t+1}), \tag{2}$$

where $A_{t+1}$ is the updated sequence of expressive robot actions. The updated action sequence replaces the remaining actions in the previously generated plan, allowing the robot to adapt its interaction behavior in response to new user instructions or changes in the environment.

### C. *Expressive Robot Controller (ERC)*

The Expressive Robot Controller executes the expressive action sequence $A$ generated by the planner utilizing a learned vision-language-action policy. Given the expressive action $a'_k$, generated by the planner, the controller produces low-level action sequence using the learned VLA $\pi_{0.5}$ [30] policy:

$$(a_t, \dots, a_{t+T_K}) = \pi(x_t, a'_k), \tag{3}$$

where $\pi(\cdot)$ denotes the VLA policy that maps the current robot state $x_t$ and expressive action $a'_k$ to low-level control commands. The generated control command $a_t$ produces coordinated motion of the mobile base and manipulator, resulting in the state transition $(p_{t+1}^{base}, q_{t+1}^{arm}) = f(p_t^{base}, q_t^{arm}, a_t)$, where $f(\cdot)$ represents the kinematic transition of the mobile base and manipulator.

## IV. MOBILE MANIPULATOR DEMONSTRATION STUDY

We investigated how observers perceived and interpreted the behaviors of an expressive mobile manipulator robot during a live demonstration study. A live demonstration was used as the study aimed to evaluate embodied interaction factors, including the robot's physical presence, motion execution, timing, and proximity to a human user [31].

| Construct | Questions | $\bar{x}$ | IQR | Min | Max |
|---|---|---|---|---|---|
| Perceived Usefulness | S1. The robot would be useful during collaborative tasks. | 4 | 2 | 1 | 5 |
| | S2. I, Personally, would be willing to use a robot like this in a collaborative task. | 3 | 2 | 2 | 5 |
| Perceived Ease of Understanding / Clarity | S3. The robot's actions were easy to understand. | 4 | 1 | 2 | 5 |
| | S4. The robot's responses were clear. | 4 | 1 | 2 | 5 |
| | S5. I could easily interpret the robot's intentions. | 4 | 1.5 | 1 | 5 |
| | S6. The robot's expressions (e.g., nodding, confused, attentive) helped clarify its behaviors. | 4 | 1 | 2 | 5 |
| | S7. The robot clearly communicated when it did not understand a request. | 4 | 1 | 2 | 5 |
| Social Expressiveness & Appropriateness | S8. The robot was socially expressive during the interaction. | 4 | 1 | 1 | 5 |
| | S9. The robot used specific social expressions (nodding, shaking, etc.) appropriately. | 4 | 1 | 2 | 5 |
| | S10. The robots' verbal responses felt socially appropriate. | 4 | 0 | 1 | 5 |
| | S11. The robot's non-verbal behaviors made the interaction more natural. | 3 | 2 | 1 | 5 |
| | S12. The robot appeared attentive when the user was speaking. | 3 | 1.5 | 1 | 5 |
| | S13. The robot waited to be engaged and only responded to the user when they engaged the robot in interactions. | 4 | 1 | 2 | 5 |
| | S14. The robot's non-verbal behaviors were used appropriately. | 3 | 2 | 1 | 4 |
| Trust, Safety & Comfort | S15. I would feel comfortable working near this robot. | 3 | 1 | 1 | 4 |
| | S16. I would feel comfortable interacting with this robot. | 3 | 2 | 1 | 4 |
| | S17. I would not trust the robot to safely hold objects. | 4 | 1.5 | 1 | 5 |
| | S18. I felt the robot did not behave in a predicable manner. | 4 | 0.5 | 2 | 5 |
| | S19. The robot appeared safe during the interaction. | 3 | 1 | 1 | 4 |

**Fig. 4.** Questionnaire for the Demonstration Study.

The objectives of the study were to determine the participants' perceptions of the ExpressMM framework in terms of: 1) perceived usefulness (**O1**), 2) behavioral clarity (**O2**), 3) social expressiveness (**O3**), and 4) perceived safety during interaction (**O4**). Prior to the demonstration, participants completed a consent form and were provided the questionnaire to complete after the demonstration. Ethics approval was obtained by the university ethics committee.

A Turtlebot-4 robot with a SO-101 manipulator arm performed a series of simple assistive tasks while interacting with a user in front of the students, Fig. 3. The tasks included: 1) following a user, 2) holding and returning handheld objects, 3) responding to spoken requests, and 4) providing spoken task guidance. The robot communicated through speech, expressive gestures, and manipulation behaviors, enabling observers to evaluate both the robot's functional capabilities and its social interaction behaviors. The video of the interaction can be found here on the YouTube link: https://youtu.be/iQOgSNQzFeI.

### A. *Participants*

The live demonstration took place during an upper year undergraduate mechatronics course. This course covers topics such as sensing, control, robot learning, and HRI, making it a natural choice for evaluating a robotic system designed to perform interactive assistive behaviors.

Twenty-three students ranged in age from 21 to 29 years $(\mu = 22.83, \sigma = 1.56)$ completed the questionnaire. Twenty-one participants identified as men and two identified as women. This gender distribution is consistent with typical engineering course enrollments [32], [33].

### B. *Measures*

A post-demonstration 5-point Likert questionnaire was used to evaluate the participants' perceptions of the robot's behaviors and interaction characteristics, Fig. 4. The questionnaire consisted of questions adapted from the Almere model [34], and the Partner Model Questionaire (PMQ) [35]. The questionnaire was organized into four constructs: 1) *Perceived Usefulness* (S1-S2) to evaluate whether participants perceived the robot's behaviors as helpful and beneficial for accomplishing the demonstrated task, a factor commonly associated with user acceptance in HRI studies

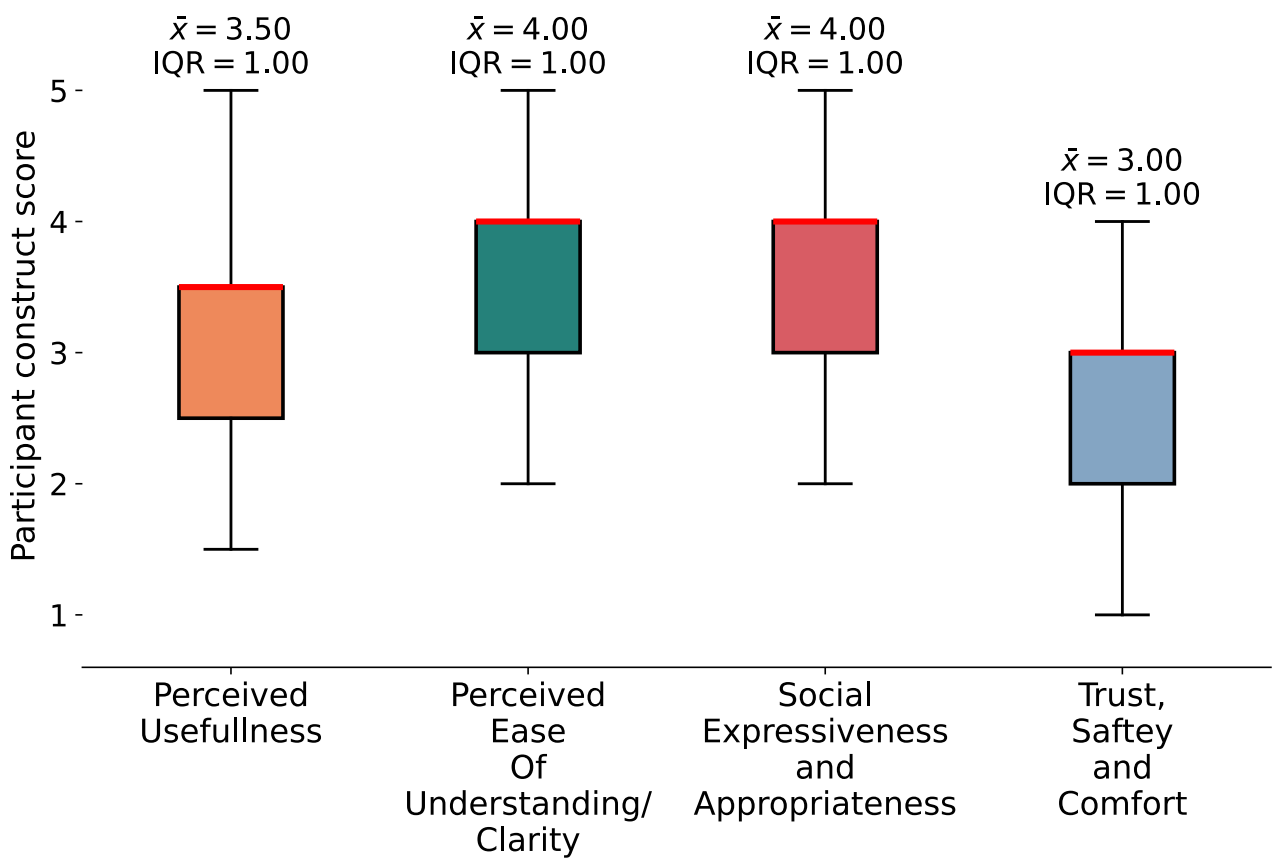

**Fig. 5.** Box-and-whisker plots for the four constructs.

[36]; 2) *Perceived Ease of Understanding / Clarity* (S3-S7), to assess how clearly the robot communicated its intentions using verbal and behavioral cues during the interaction; 3) *Social Expressiveness and Appropriateness* (S8-S14), to evaluate the robot's ability to convey socially meaningful cues through its behaviors and if they were socially appropriate; and 4) *Trust, Safety, and Comfort* (S15-S19), evaluates participants' perceptions of the robot's behavioral predictability, perceived safety, and overall comfort during interaction.

## C. *Procedure*

The study was conducted during a regular class session in which participants observed the live demonstration of the mobile manipulator robot interacting with one of the researchers. The purpose of the demonstration was to illustrate how the robot performs collaborative assistive tasks while communicating its intentions through speech and expressive behaviors.

Two collaborative interaction scenarios were demonstrated: 1) in the first scenario, the robot followed the user through a workspace and assisted by holding an object before returning it to the user upon request. The robot also responded to ambiguous or unclear commands by expressing confusion or declining unsafe actions; and 2) in the second scenario, the robot assisted during a simple assembly task involving the installation of rods on a TurtleBot platform. In this scenario, the robot held a tool for the user, returned it when requested, and provided verbal guidance when the user encountered a mechanical issue during assembly. These interactions demonstrated how the robot integrates task-oriented assistance with expressive behaviors such as attentiveness, acknowledgment gestures, confusion signals, and celebratory responses.

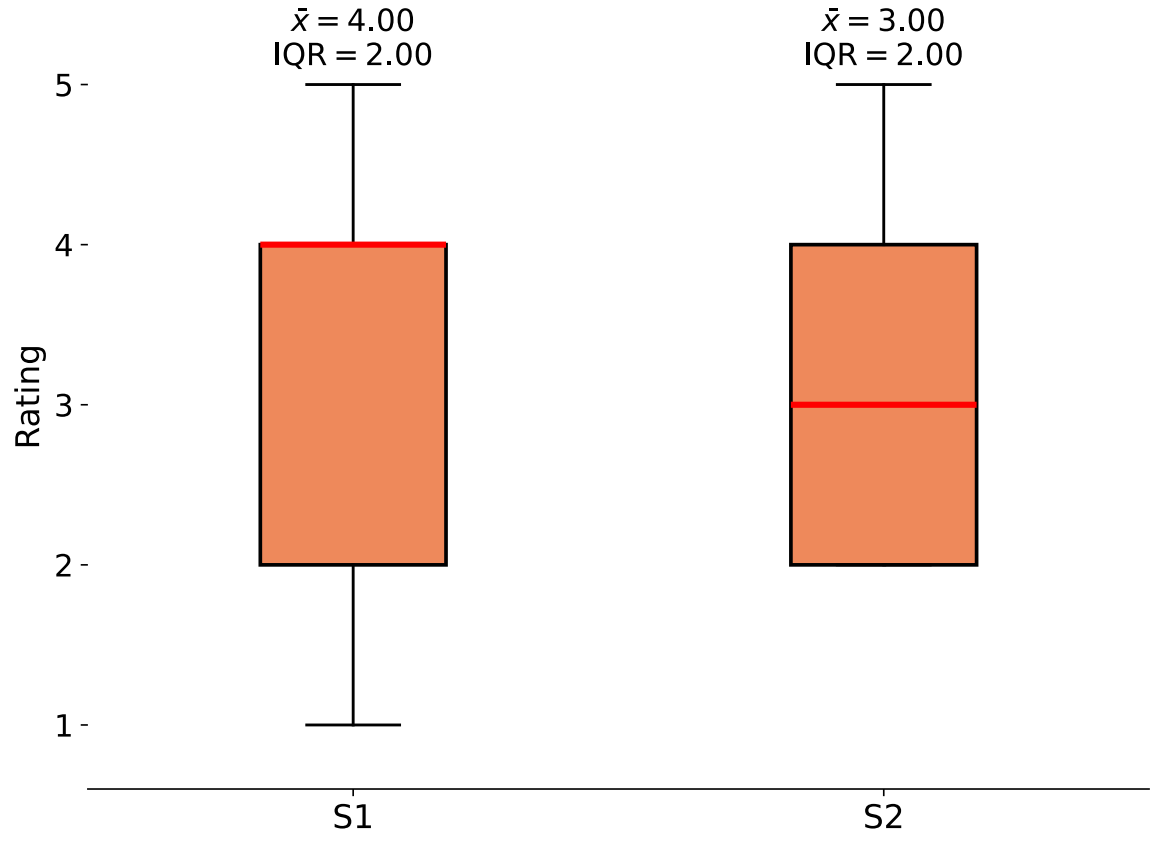

**Fig. 6.** Box-and-whisker plots for the *Percieved Usefulness*.

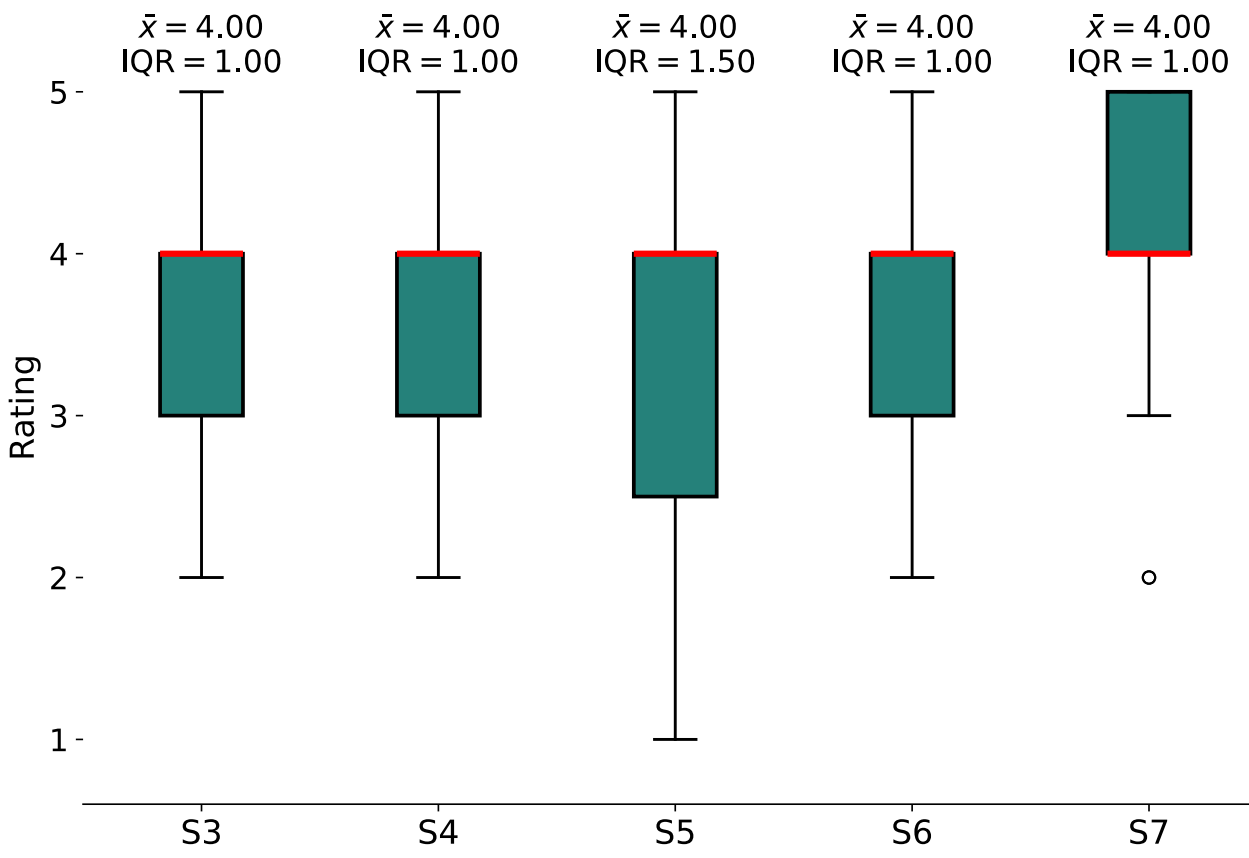

**Fig. 7.** Box-and-whisker plots for the *Perceived Ease of Understanding / Clarity* construct.

# V. RESULTS

We conducted a series of Shapiro-Wilk tests for normality [37] to evaluate the distribution of responses for each questionnaire item. The results indicated that the majority of questionnaire items significantly deviated from normality ($p < 0.05$). Therefore, non-parametric statistical tests were used to interpret the results.

Cronbach's alpha [38] was used to determine the reliability of each construct. The reliability analysis showed acceptable values across the constructs with respect to exploratory HRI studies [39]: 1) *Perceived Usefulness*, $\alpha = 0.728$ ; 2) *Perceived Ease of Understanding / Clarity*, $\alpha = 0.721$; 3) *Social Expressiveness and Appropriateness*, $\alpha = 0.710$; and 4) *Trust, Safety, and Comfort*, $\alpha = 0.658$.

Fig. 5 presents the box and whisker plots for the constructs across all participants. The *Perceived Usefulness* construct had a median score of $\bar{x} = 3.5$ with $IQR = 1$. Both the *Perceived Ease of Understanding / Clarity* construct and the *Social Expressiveness and Appropriateness* construct had $\bar{x} = 4$ with $IQR = 1$. Finally, the *Trust, Safety and Comfort* construct had a $\bar{x} = 3$ with $IQR = 1$.

## A. *Perceived Usefulness*

The box and whisker plots for the *Perceived Usefulness construct* are presented in Fig. 6. Participants rated S1 positive ($\bar{x} = 4, IQR = 2$) and S2 as neutral ($\bar{x} = 3, IQR = 2$).

## B. *Perceived Ease of Understanding / Clarity*

The box and whisker plots for the *Perceived Ease of Understanding / Clarity* construct are presented in Fig. 7. Participants rated the statements S3 and S4 the same ($\bar{x} = 4$, $IQR = 1$). Participants also agreed with the statements S5 ($\bar{x} = 4, IQR = 1.5$), S6 ($\bar{x} = 4, IQR = 1$) and S7 ($\bar{x} = 4$, $IQR = 1$).

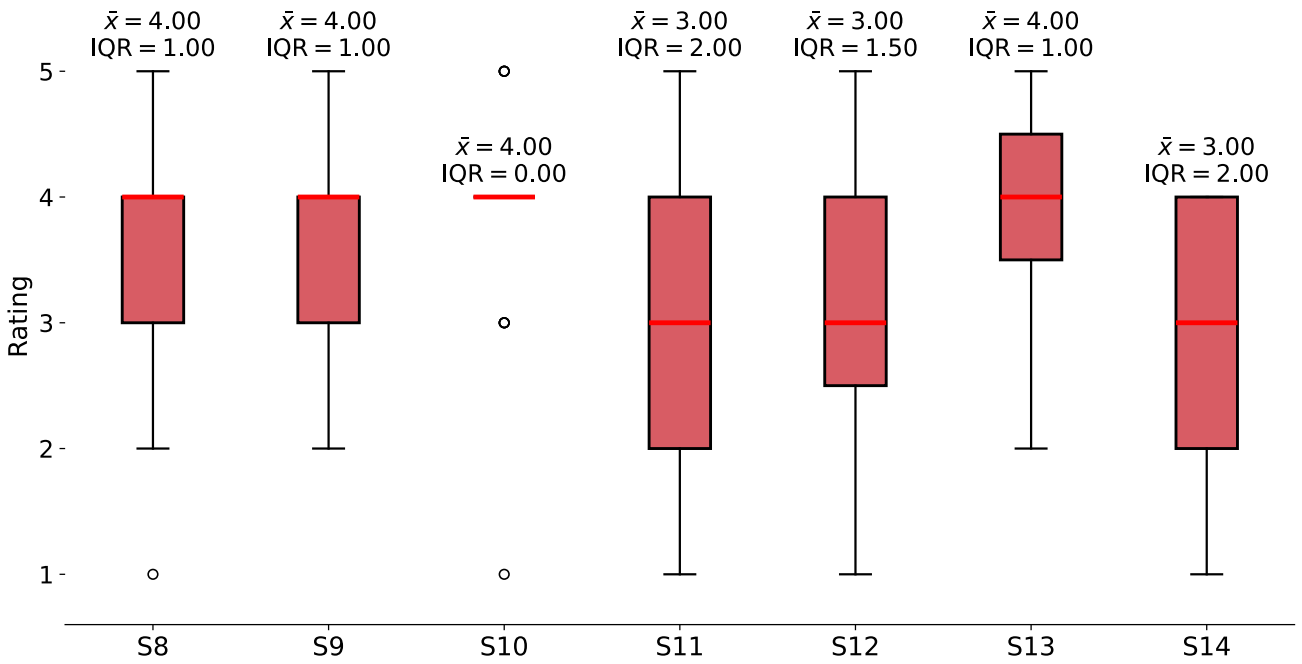

**Fig. 8.** Box-and-whisker plots for the *Social Expressiveness and Appropriateness* construct.

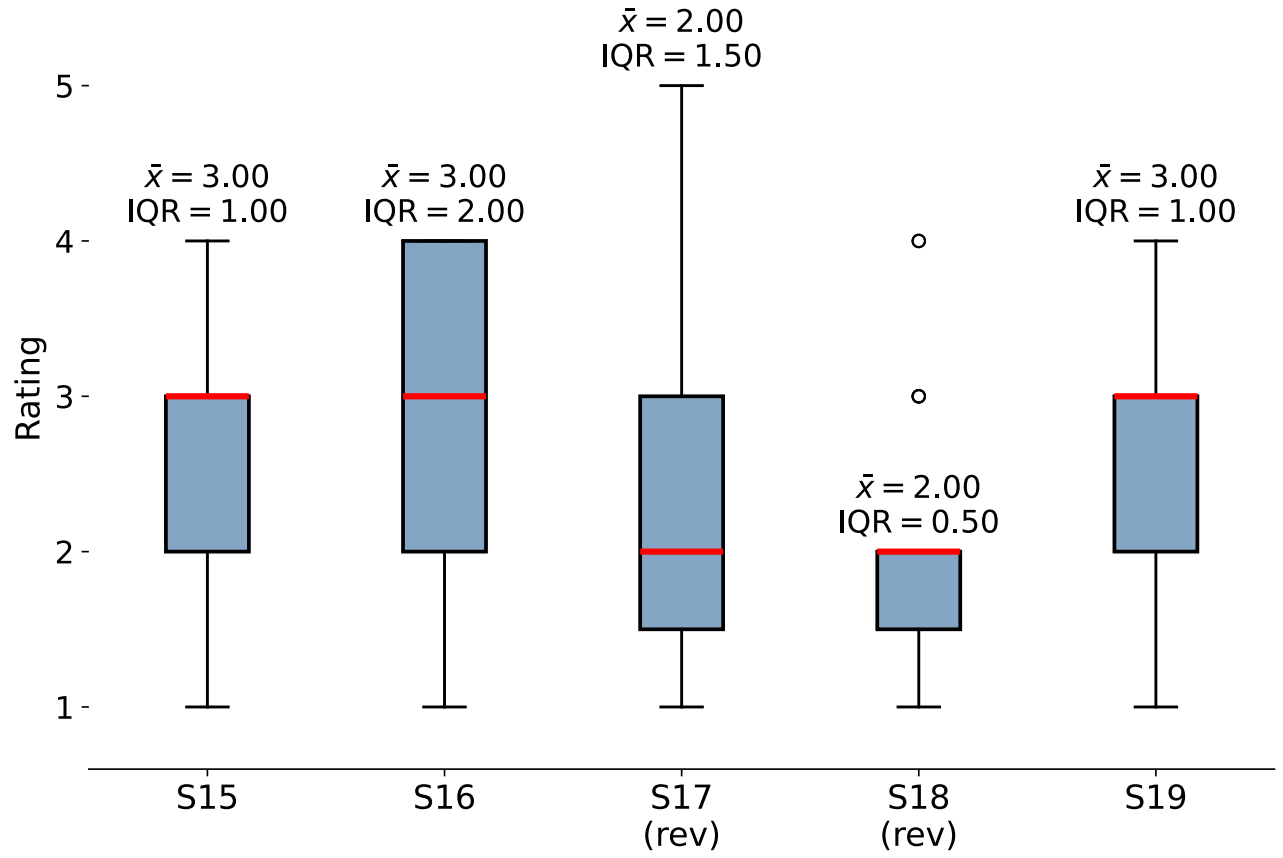

**Fig. 9.** Box-and-whisker plots for the *Trust, Safety, and Comfort* construct.

### C. *Social Expressiveness and Appropriateness*

The box and whisker plots for the *Social Expressiveness and Appropriateness* construct are presented in Fig. 8. Participants positively rated the statements S8, S9 and S13 ($\bar{x} = 4, IQR = 1$). The statement S10 was rated positively with no spread in the interquartile range ($\bar{x} = 4$, $IQR = 0$). However, more neutral ratings and greater response variability were found for statements S11 and S14 ($\bar{x} = 3$, $IQR = 2$), with less variability for S12 ($\bar{x} = 3$, $IQR = 1.5$).

### D. *Trust, Safety, and Comfort*

The box and whisker plots for the *Trust, Safety, and Comfort* construct are presented in Fig. 9. Participants rated the statements S15 and S16 as neutral ($\bar{x} = 3, IQR = 1$; $\bar{x} = 3, IQR = 2$). For the reverse-coded trust statement S17, particpants had some reservations ($\bar{x} = 4, IQR = 1.5$). Participants felt the robot behaved in a preditable manner with the reverse-coded statement S18 with responses showing minimal spread ($\bar{x} = 4, IQR = 0.5$). Finally, participants rated S19 as neutral ($\bar{x} = 3, IQR = 1$).

## VI. DISCUSSIONS

This live demo study investigated observer perceptions of an expressive mobile manipulator incorporated with our ExpressMM framework. For **O1**, participants perceived the robot as useful during the demonstrated collaborative tasks, although responses were more mixed regarding whether they would personally rely on it. This is consistent with prior HRI research showing that positive perceptions of robot assistance do not necessarily translate into full user willingness to use a robot during collaborative tasks [40]. For **O2**, participants rated the robot's actions, responses, and intentions as easy to understand, indicating positive perceptions of behavioral clarity. Prior HRI research has also shown that robot verbal and non-verbal cues can help people interpret robot intent during interactions [41]. Overall, participants perceived the robot's behaviors as expressive and appropriate (**O3**). However, some of its non-verbal expressions such as waiting for the user during the interaction, and appearing attentive while the user was speaking received mix ratings. These results are consistent with previous HRI studies that have shown that robot social and nonverbal cues influence how people interpret interaction quality and social meaning during HRI [41], [42]. Participants were cautious in their ratings of trust, safety, and comfort (**O4**), with largely neutral responses for comfort and safety while having reservations about trusting the robot to safely hold objects. This aligns with HRI studies that have shown that trust and perceived safety can be lower in physical HRI involving object handling near users [3].

As this is an exploratory study with a single engineering classroom cohort in a live demonstration setting, these findings may not extend to other populations, application contexts, or longer-term interactions. Our sample was gender imbalanced, and future studies should include larger and more demographically balanced groups, as prior HRI research has shown that user responses to robots vary across gender [43].

## VII. CONCLUSIONS

We developed the novel ExpressMM architecture which generates expressive behaviors for a mobile manipulator during collaborative HRI tasks. Our live demonstration HRI study showed that ExpressMM can faciliate expressive, functional and interruptible whole-robot-body HRI, where users can provide corrective or redirecting instructions to the robot during task execution. Overall, in the live demonstration, participants perceived the robot as clear, socially expressive, and safe to interact with. These results suggest that integrating expressive behaviors with task-oriented mobile manipulation can support clear intent communication and understandable robot actions during collaboration.

## ACKNOWLEDGEMENTS

The authors would like to thank Glenn Takashi Shimoda, Sujith Santharuban, Yaseen Mohammad Rehman, Abigail Nevo and Georgia Jovanovic from the ASBLab at the University of Toronto for their assistance with the experimental setup, video and photo demonstrations.